# MULTI-VIEW TASK-DRIVEN RECOGNITION IN VISUAL SENSOR NETWORKS

*Ali Taalimi, Alireza Rahimpour, Liu Liu and Hairong Qi*

Department of Electrical Engineering and Computer Science
University of Tennessee, Knoxville, TN, USA 37996
{ataalimi, arahimpo, lliu25, hqi}@utk.edu

## ABSTRACT

Nowadays, distributed smart cameras are deployed for a wide set of tasks in several application scenarios, ranging from object recognition, image retrieval, and forensic applications. Due to limited bandwidth in distributed systems, efficient coding of local visual features has in fact been an active topic of research. In this paper, we propose a novel approach to obtain a compact representation of high-dimensional visual data using sensor fusion techniques. We convert the problem of visual analysis in resource-limited scenarios to a multi-view representation learning, and we show that the key to finding properly compressed representation is to exploit the position of cameras with respect to each other as a norm-based regularization in the particular signal representation of sparse coding. Learning the representation of each camera is viewed as an individual task and a multi-task learning with joint sparsity for all nodes is employed. The proposed representation learning scheme is referred to as the multi-view task-driven learning for visual sensor network (MT-VSN). We demonstrate that MT-VSN outperforms state-of-the-art in various surveillance recognition tasks.

## 1. INTRODUCTION

Distributed camera networks is an important cross-disciplinary research field emerged from computer vision, distributed processing, embedded computing and sensor networks. Computer vision tasks require high-end computational power as well as memory, that is why, traditionally, most computer vision systems have been implemented on workstations. However, networks of distributed smart cameras can solve computer vision problems by providing valuable information through distributed sensing and multi-view processing. Though, a major challenge in visual sensor networks is the limitation in terms of transmission bandwidth.

Early studies in distributed visual recognition systems acquired images and compressed locally at the camera nodes, and then transmitted to the base station which performs the specific analysis tasks. In some recent approaches [1, 2, 3, 4], the visual features (*e.g.* SIFT) have been extracted locally in each camera and then compressed and transmitted to the base station for further analysis (*e.g.* classification).

Sparse models have been successfully applied in the recent two decades in many scientific disciplines [5, 6, 7, 8, 9, 10]: sparsity principle in statistics and machine learning selects a simple model out of a pool of possible choices, which are called atoms or elements. The set of all the atoms is called dictionary, and the coefficients of the linear combinations are called sparse codes. Our method learns an offline dictionary for each view to converting the physical pixel-based information into the latent space of sparse codes. In this scenario, each camera produces and transmits a compressed discriminative representation that is band-limited and memory-efficient. The final recognition is done in the base station by applying view-dependent classifiers in a classifier fusion scheme. Each classifier is learned simultaneously with its corresponding view dictionary in a task-driven learning scheme. We will demonstrate that by obtaining each view-dependent dictionary and classifier jointly in a fully supervised setting that is tuned for the prediction task, our method can generate a discriminative representation in each view, and multi-view standpoint while at the same time to be efficiently compact, suitable for memory and pro- cess limited systems like visual sensor networks.

Any source of information is limited to its neighborhood [11, 12]. Making decision relying on a single source can jeopardize the decision-making process [3, 13, 14, 15]. One solution is information fusion from different sensors that has been demonstrated to be more robust to sensor failure [16]. For instance, each camera is considered as a modality in [17] and feature-fusion techniques is applied which led to a superior multi-view action recognition.

Fusion can be conducted in either the feature level or classifier level [18, 19]. In feature-fusion, different types of features are combined to make a new feature set while classifier-fusion aggregates decisions from several classifiers which are individually trained on various features. Although feature-fusion has been more efficient, the algorithm design is challenging especially when the features have different dimensions, and consequently, it is a relatively less-studied prob-lem [18]. The simplest way of feature-fusion is to concatenate features into one high-dimensional vector. Besides higher dimension, the concatenated feature vector also does not contain the valuable information of correlation between feature types. In this paper, we use "view", "sensor", and "modality" interchangeably.

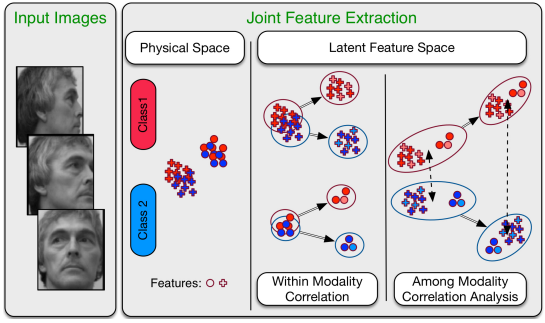

**Fig. 1**. Illustration of the proposed dictionary learning framework for multi-view face recognition on UMIST [20]. Consider each person has $M = 3$ different poses (modalities) from profile to frontal views. Originally, the physical attributes are not discriminative enough to identify the target. Our method learns a set of dictionaries to find the representation of data in latent space of sparse codes, to make classes more distinctive in each view, and, from multi-view standpoint.

We propose a method that is designed to learn a proper representation of multi-view signals for the specific task of distributed recognition. Our approach translates physical representation of the signal from various views, into a latent feature space. As Fig. (1) shows, in the latent space, classes can be easily separated using a compact and discriminative representation. We encode correlation between view-modalities through a norm-based regularization in the particular signal representation of sparse coding. The feature-fusion in physical space is translated as grouping in the space of sparse codes. A supervised dictionary learning is proposed to obtain a reconstructive and discriminative dictionary with a small number of atoms for each feature that can generate discriminative latent feature in the space of sparse codes (Fig. 1).

**Notation**. Suppose a set of $N$ training input and output pairs denoted by $\{(\boldsymbol{X}^i, \boldsymbol{l}^i)\}_{i=1}^N$, where the $i$-th sample $\boldsymbol{X}^i$ is a multimodal data that is seen from $M$ modalities: $\boldsymbol{X}^i = \{\boldsymbol{x}_1^i, \ldots, \boldsymbol{x}_M^i\}$, where $\boldsymbol{x}_m^i$ is a vector in $\mathbf{R}^{n_m}$ with $n_m$ being the dimension of the $m$-th modality. We use one-hot encoding to denote the label of the $i$-th sample as binary vector $\boldsymbol{l}^i$. We consider the dictionary of the $m$-th modality with $p$ elements in $\mathbf{R}^{n_m \times p}$ as $\boldsymbol{D}_m = [\boldsymbol{d}_m^1, \ldots, \boldsymbol{d}_m^p]$ where each element or atom $\boldsymbol{d}_m$ is a vector of size $n_m$. That is, the $r$-th dictionary element (atom) is a multimodal feature from $M$ modalities, $\{\boldsymbol{d}_1^r, \ldots, \boldsymbol{d}_M^r\}$. For simplicity, we show the set of all dictionaries as $\{\boldsymbol{D}_m\}$.

## 2. BACKGROUND

**Single View Dictionary Learning.** Since this step is same for all modalities, we omit index $m$ in this section for simplicity. The classical dictionary learning estimates the dictionary $\boldsymbol{D} = [\boldsymbol{d}^1, \ldots, \boldsymbol{d}^p]$ with $p$ atoms, from the data using an unsupervised loss function [21, 22]. The generated sparse codes is used as a latent feature vector of data $\boldsymbol{x}$ for predicting the its label $\boldsymbol{l}$ by training a classifier in the classical expected risk optimization [23]. However, this approach is sub-optimal because dictionary is fixed during training classifier and is independent of the labels. So, the dictionary learning does not fully utilize the label information.

LC-KSVD [24] utilizes the class labels information to generate a discriminative and compact dictionary in an all-vs-all scheme, where the dictionary is shared between classes. In order to learn dictionaries with uncorrelated atoms, LC-KSVD forces each atom to represent only one class. Assuming $i$-th training sample $\boldsymbol{x}^i$ from the $c$-th class, a binary vector $\boldsymbol{q}^i \in \mathbf{R}^p$ is define that is zero everywhere except at the indices of atoms which belong to the $c$-th class. This so called "label consistency constraint" is applied using $\{\boldsymbol{q}_i\}_{i=1}^N$ so that the sample from $c$-th class is represented using the same subset of dictionary items associated with class $c$:

$$\operatorname*{argmin}_{\boldsymbol{D},\boldsymbol{T},\boldsymbol{W},\boldsymbol{x}^i} \sum_{i=1}^N \|\boldsymbol{x}^i - \boldsymbol{D}\boldsymbol{\alpha}^i\|_{\ell_2}^2 + \alpha \|\boldsymbol{q}^i - \boldsymbol{T}\boldsymbol{\alpha}^i\|_{\ell_2}^2 +$$
$$\beta \|\boldsymbol{l}^i - \boldsymbol{W}\boldsymbol{\alpha}^i\|_{\ell_2}^2 + \lambda \|\boldsymbol{\alpha}^i\|_{\ell_1} \quad (1)$$

where $\boldsymbol{T}$ is a linear transformation matrix, $\boldsymbol{W}$ is the parameters of a linear classifier and $\alpha$ and $\beta$ are regularization parameters of label consistency and miss-classification error, respectively. The label consistency $\|\boldsymbol{q}^i - \boldsymbol{T}\boldsymbol{\alpha}^i\|_{\ell_2}^2$ regularization enforces the linear transformed version of original sparse codes $\boldsymbol{T}\boldsymbol{\alpha}^i$ to be most discriminative in the $\mathbf{R}^p$ space.

**Issues.** The typical way to solve problems like (1) is stochastic gradient descent. The gradients of the objective function with respect to $\boldsymbol{D}$ and $\boldsymbol{W}$ are $\partial/\partial \boldsymbol{D} = -(\boldsymbol{x}^i - \boldsymbol{D}\boldsymbol{\alpha}^i)\boldsymbol{\alpha}^{i\top}$ and $\partial/\partial \boldsymbol{W} = -\beta(\boldsymbol{l}^i - \boldsymbol{W}\boldsymbol{\alpha}^i)\boldsymbol{\alpha}^{i\top}$, respectively. It is evident that $\partial/\partial \boldsymbol{D}$ does not have any effect from $\boldsymbol{W}$ and vice versa. Hence, each variable is updated independent of the other variable. We want to make the connection between the dictionary $\boldsymbol{D}$ and the classifier $\boldsymbol{W}$ in each view (modality) so that the incremental update of each optimization variable has access to the information of other variables, in a gradient descent fashion. This way, we design our method to learn a dictionary that is adapted to the specific task of multi-view object recognition in a task-driven learning scheme.

Also, we want to generalize problem (1) to be able to fuse information at the feature level. The idea is to estimate a set of dictionaries and classifiers for $M$ modalities so that in each modality the dictionary $\boldsymbol{D}_m$ can reconstruct the signal $\boldsymbol{x}_m^i$ with the sparse coefficient vector $\boldsymbol{\alpha}_m^{i\star}$ that is distinctive enough so that a simple linear classifier with parameters $\boldsymbol{W}_m$ can estimate the label of the signal in the base station, *e.g.* $l^i = W_m \alpha_m^{i\star}$. Next, we explain in more details our approach to solving the above limitations.

### 2.1. Multi-View Task Driven Dictionary Learning

The dictionary $D_m$ decomposes data in $m$-th modality, $x_m^i$, to a sparse coefficient vector $\alpha_m^{i\star} \in \mathbf{R}^p$. We consider $A^{i\star}$ in $\mathbf{R}^{p \times M}$ as matrix of multimodal sparse codes of $\{x_m^i\}_{m=1}^M$ which is made by horizontally concatenating the sparse codes from all modalities: $A^{i\star} = [\alpha_1^{i\star}, \ldots, \alpha_M^{i\star}]$. Given $N$ training samples $\{x_m^i\}_{i=1}^N$ and $m$ in $\{1, \ldots, M\}$ with their label vector $\{l^i\}_{i=1}^N$, we propose to obtain jointly, the multimodal sparse representation $A^i$, a set of task-driven dictionaries $\{D_m\}_{m=1}^M$, and classifiers $\{W_m\}_{m=1}^M$ in a bi-level optimization.

For each multimodal input $\{x_m^i\}_{m=1}^M$, we obtain the multimodal sparse representation $A^{i\star}$ in $\mathbf{R}^{p \times M}$ using the inner optimization problem which is designed to enforce collaboration between different modalities:

$$\underset{A^i}{\operatorname{argmin}} \sum_{m=1}^M \frac{1}{2} \|x_m^i - D_m \alpha_m^i\|_{\ell_2}^2 + \alpha \|q^i - T_m \alpha_m^i\|_{\ell_2}^2 + \lambda_1 \|A^i\|_{\ell_{12}} + \frac{\lambda_2}{2} \|A^i\|_F^2 \quad (2)$$

where $\lambda_1$ and $\lambda_2$ are the regularization parameter and $\|.\|_F$ is the Frobenius norm. When $\lambda_2 > 0$ problem (2) is a generalization of elastic-net optimization [21] and it has been proved in [25, 17] that it leads to more stable results.

Fusion in physical space would be grouping in space of sparse codes and is enforced by $\|A^i\|_{\ell_{12}} = \sum_{r=1}^p \|A_{r\rightarrow}\|_2$; where $A_{r\rightarrow}$ is the $r$-th row in $A^i$. The $\|A^i\|_{\ell_{12}}$ promotes a solution with sparse non-zero rows. Hence, similar support is enforced on $A^i$ at the column level of each dictionary $D_m$ [26].

The outer-level objective is designed to jointly estimate $\{D_m, W_m, T_m\}$, the dictionary, classifier and the transformation in modality $m$:

$$\underset{W_m, D_m}{\operatorname{argmin}} \sum_{i=1}^N \frac{1}{2} \|l^i - W_m \alpha_m^{i\star}(x_m^i, D_m)\|_{\ell_2}^2 + \frac{\nu}{2} \|W_m\|_F^2 \quad (3)$$

where $\nu = 0.1$ is the regularization parameter, and $\alpha_m^{i\star}(x_m^i, D_m)$ is the solution of inner-level Eq. (2). The first term in Eq. (3) is the supervised convex loss function that evaluates how close classifier with parameters $W_m$ using $\alpha^{i\star}(x_m^i, D_m)$ can predict label $l^i$. The multimodal sparse coefficients, $A^{i\star}$ is a parameter for outer-level, but variable for the inner-level objective. Also, the dictionary $D_m$ is not explicitly defined in optimization problem (3); but, it is defined implicitly in inner-level problem (2).

**Optimization.** The problem (2) has the product of the two optimization variables as $D\alpha_m^i$; which implies that this problem is not joint convex in the space of coefficients and dictionary. However, when one of the two optimization variables is fixed, the problem (2) is convex with respect to the other variable [5]. Hence, the bi-level optimization is solved by splitting to two sub-problems, and in each sub-problem, we solve the optimization for one variable while others are fixed. We solve the proposed bi-level optimization problem (2) and (3) following [17, 26] using the SPArse Modeling Software [27].

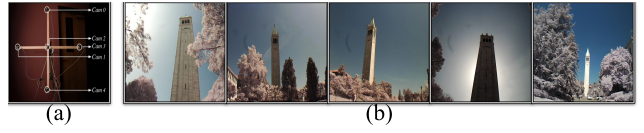

**Fig. 2**. (a) Apparatus which instruments five camera sensors [28]. (b) Large-baseline images from different vantage points

### 2.2. Classification Approach

The output of training phase would be the dictionary and classifier in each modality, $\{D_m, W_m\}_{m=1}^M$. Each test sample $X^t$ is observed from the same set of $M$ modalities, $X^t = \{x_m^t\}_{m=1}^M$. We obtain the corresponding multimodal sparse codes, $\{\alpha_m^t\}_{m=1}^M$, by solving Eq. (2), given $\{x_m^t\}$ and $\{D_m\}$. The query is assigned to the class with maximum summation of classification scores of all modalities in majority voting scheme, $\operatorname{argmax} \sum_{m=1}^M W_m \alpha_m^t$.

### 3. EXPERIMENTS

In this section we evaluate the performance of MT-VSN in two different applications: face recognition in surveillance video using the UMIST dataset [20] and distributed object recognition in smart camera networks using the Berkeley Multiview Wireless (BMW) [28]. Samples are normalized to have zero mean and unit $\ell_2$ norm. To compare with the performance of unimodal dictionary learning algorithms, we learn independent dictionaries and classifiers for each modality and then combine the individual scores for a fused decision. It means, the joint sparsity regularization $\ell_{12}$-norm in in Eq. (2) is replaced by specificity regularization [29], $\|A\|_{\ell_{11}} = \sum |A_{ij}|$. This is equivalent to decision fusion where the final decision is achieved by aggregating the individual scores from each modality. The $\ell_{11}$ does not encode feature-fusion, because $\ell_1$-norm is blind to see the relation between variables [30]. The hyperparameters $\lambda_1$ and $\lambda_2$ in Eq. (2) are selected from the set $\{0.0005, 0.001, 0.005, 0.01, 0.05\}$ by cross validation, and $\alpha = 4.0$ and $\beta = 2.0$, as LC-KSVD [24].

**Distributed Object Recognition**. The BMW consists of multiple-view images of 20 landmark buildings on the campus of University of California, Berkeley. The images are taken by five cameras with four of them located on the pe-riphery of the cross and one in the center. It provides 16 images (vantage points) for each building as shown in Fig. 2. The dataset is shipped with three feature modalities for each image: sift, surf and chog [31]. Following [2] training phase in Sec.(2.1) is done using 8 images from even vantage points of the central camera. Evaluation is done on the other cameras. Same as [28], we evaluate the recognition performance using one camera (i.e., Cam2) and two cameras (i.e., Cam1 and Cam2). We report the performance of the state-of-the-art dictionary learning methods LC-KSVD [24], TDL [25] and sparse PCA [28] and MT-VSN in Table (1).

Table 1. The recognition rate on large-baseline evaluation of BMW.

| BMW% | MT-VSN | | | | LC-KSVD [24] | | | | sPCA [28] | | | | TDL [25] | | | |
|---|---|---|---|---|---|---|---|---|---|---|---|---|---|---|---|---|
| | siftsurf | sifthog | surfhog | $\ell_{12}$ | sift | surf | hog | $\ell_{11}$ | sift | surf | hog | $\ell_{11}$ | sift | surf | hog | $\ell_{11}$ |
| **1 Cam** | 94.26 | 94.20 | 95.11 | **96.73** | 89.02 | 90.04 | 90.62 | 92.23 | 71.25 | 80.62 | 81.88 | 84.37 | 92.75 | 92.26 | 93.35 | 94.72 |
| **2 Cam** | 97.33 | 97.95 | 98.06 | **99.14** | 91.08 | 91.73 | 92.02 | 94.64 | 76.88 | 88.13 | 93.75 | 94.58 | 93.25 | 93.25 | 94.68 | 95.58 |

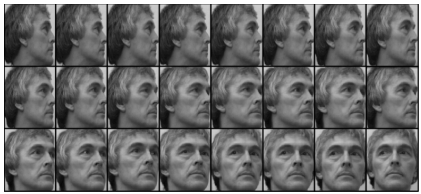

**Fig. 3**. Different poses of a subject from UMIST database. Each row is a view-range or modality for the subject.

Table 2. Multi-view face recognition on UMIST datasets

| | JDSRC [32] | MTSRC [30] | MCWDL [26] | **MT-VSN** |
|---|---|---|---|---|
| 2 Views | 86.52 | 88.42 | 91.12 | **93.30** |
| 3 Views | 98.96 | 99.63 | 99.20 | **100.0** |

We assign 8 atoms per class that leads to $p = 160$ atoms in all the settings. We report performance of other methods for single feature (sift, surf or chog) and for multimodal setting under the $\ell_{11}$, which is equal to decision fusion. We report MT-VSN when there are two features available (siftsurf, sifthog and surfhog) and when all three are available under $\ell_{12}$. Note that, we did not report MT-VSN for one feature because, it exploits correlation between different sources and when only one source is available it is similar to [24, 25]. Table 1 testifies on the superiority of feature-fusion on decision-fusion. That is, our method is designed to efficiently represent multimodal data by exploiting the relation (correlation) between modalities, while other methods are blind to see the coupling between variables and treat them independently.

**Multi-view Face Recognition in Surveillance Video.** The UMIST face database consists of 564 cropped images of 20 persons with mixed race and gender [20]. Each person has different poses from profile to frontal views. The setup is unconstrained and faces may have pose variations within each view-range. We run multi-view face recognition on UMIST by segmenting views of each person to $M$ different view-ranges with equal number of images. Intensity values were used. In Fig. (3), the poses of a subject are divided in $M = 3$ view-ranges. We report the performance of the MT-VSN for 2 and 3 views. Table (2) has the the results of 10-fold cross validation. The corresponding dictionary of each view has one normalized image from each subject in that view, $p = 20$. MT-VSN learns a dictionary with uncorrelated class-specific atoms and outperforms [30, 32] by more than 5% and enhances [26] more than 2%.

## 4. CONCLUSION

We presented a new method for learning multimodal dictionaries while its sparse representations share the same sparsity pattern at the atom level of modality-based dictionaries. We convert the problem of reducing the dimensionality of visual features in distributed systems, to the multimodal representation learning. In the context of multimodal data analysis, the position of cameras with respect to each other (modality configuration) induces a strong correlation structure between modalities. Our design is able to encode the modality configuration in the latent space and find the low-dimensional and discriminative representation. We show quantitatively and qualitatively that our method outperforms state-of-the-art in various distributed computer vision tasks. This is the result of carefully designed norm-base regularization that can represent the multimodal data seen from multiple cameras, and hence show promising to deal with the overfitting problem.